\documentclass[11pt,letterpaper]{article}
\usepackage{emnlp2017}
\usepackage{times}
\usepackage{latexsym}
\usepackage{epsfig}
\usepackage{amssymb}
\usepackage{amsmath}
\usepackage{amsfonts}
\usepackage{multirow}
\usepackage[linesnumbered,ruled]{algorithm2e}
\usepackage{graphicx}
\usepackage{subfigure}
\usepackage{url}
\usepackage{array}
\usepackage{tikz}
\usepackage{booktabs}
\usepackage{enumitem}

\emnlpfinalcopy

\def\emnlppaperid{313}

\title{Entity Linking for Queries by Searching Wikipedia Sentences}

\author{Chuanqi Tan$^\dag$
	\Thanks{\;Contribution during internship at Microsoft Research.}
	\hspace{0.15cm} Furu Wei$^\ddag$ \hspace{0.1cm} Pengjie Ren$^+$\hspace{0.1cm} Weifeng Lv$^\dag$ \hspace{0.1cm} Ming Zhou$^\ddag$\\
	$^\dag$State Key Laboratory of Software Development Environment, Beihang University, China \\
	$^\ddag$Microsoft Research Asia \hspace{2.5cm}
	$^+$Shandong University \\
	$^\dag${\tt tanchuanqi@nlsde.buaa.edu.cn} \hspace{0.5cm} $^+${\tt jay.ren@outlook.com} \\
	$^\ddag${\tt \{fuwei, mingzhou\}@microsoft.com} \hspace{0.5cm}	$^\dag${\tt lwf@buaa.edu.cn} 
}

\date{}

\begin{document}

	\maketitle
	
	\begin{abstract}
		
		We present a simple yet effective approach for linking entities in queries. The key idea is to search sentences similar to a query from Wikipedia articles and directly use the human-annotated entities in the similar sentences as candidate entities for the query. Then, we employ a rich set of features, such as link-probability, context-matching, word embeddings, and relatedness among candidate entities as well as their related entities, to rank the candidates under a regression based framework.	The advantages of our approach lie in two aspects, which contribute to the ranking process and final linking result. First, it can greatly reduce the number of candidate entities by filtering out irrelevant entities with the words in the query. Second, we can obtain the query sensitive prior probability in addition to the static link-probability derived from all Wikipedia articles. We conduct experiments on two benchmark datasets on entity linking for queries, namely the ERD14 dataset and the GERDAQ dataset. Experimental results show that our method outperforms state-of-the-art systems and yields 75.0\% in F1 on the ERD14 dataset and 56.9\% on the GERDAQ dataset. 
		
	\end{abstract}
	
	\section{Introduction}
	Query understanding has been an important research area in information retrieval and natural language processing~\citep{croft2010query}. A key part of this problem is entity linking, which aims to annotate the entities in the query and link them to a knowledge base such as Freebase and Wikipedia. This problem has been extensively studied over the recent years~\citep{Q15-1023,usbeck2015gerbil,cornolti2016piggyback}. 
	
	The mainstream methods of entity linking for queries can be summed up in three steps: mention detection, candidate generation, and entity disambiguation. The first step is to recognize candidate mentions in the query. The most common method to detect mentions is to search a dictionary collected by the entity alias in a knowledge base and the human-maintained information in Wikipedia (such as anchors, titles and redirects)~\citep{laclavik2014search}. The second step is to generate candidates by mapping mentions to entities. It usually uses all possible senses of detected mentions as candidates. Hereafter, we refer to these two steps of generating candidate entities as entity search. Finally, they disambiguate and prune candidate entities, which is usually implemented with a ranking framework. 	
	
	There are two main issues in entity search. First, a mention may be linked to many entities. The methods using entity search usually leverage little context information in the query. Therefore it may generate many completely irrelevant entities for the query, which brings challenges to the ranking phase. For example, the mention ``Austin'' usually represents the capital of Texas in the United States. However, it can also be linked to ``Austin, Western Australia'', ``Austin, Quebec'', ``Austin (name)'', ``Austin College'', ``Austin (song)'' and 31 other entities in the Wikipedia page of ``Austin (disambiguation)''. For the query ``blake shelton austin lyrics'', Blake Shelton is a singer and made his debut with the song ``Austin''. The entity search method detects the mention ``austin'' using the dictionary. However, while ``Austin (song)'' is most related to the context ``blake shelton'' and ``lyrics'', the mention ``austin'' may be linked to all the above entities as candidates. Therefore candidate generation with entity search generates too many candidates especially for a common anchor text with a large number of corresponding entities. Second, it is hard to recognize entities with common surface names. The common methods usually define a feature called ``link-probability'' as the probability that a mention is annotated in all documents. There is an issue with this probability being static whatever the query is. We show an example with the query ``her film''. ``Her (film)'' is a film while its surface name is usually used as a possessive pronoun. Since the static link-probability of ``her'' from all Wikipedia articles is very low, ``her'' is usually not treated as a mention linked to the entity ``Her (film)''. 
	
	In this paper, we propose a novel approach to generating candidates by searching sentences from Wikipedia articles and directly using the human-annotated entities as the candidates. Our approach can greatly reduce the number of candidate entities and obtain the query sensitive prior probability. We take the query ``blake shelton austin lyrics'' as an example. Below we show a sentence in the Wikipedia page of ``Austin (song)''.
	
	\begin{table}[h]
		\centering
		\begin{tabular}{ |p{7cm}| }
			\hline			
			\textbf{[[Austin (song)$\vert$Austin]]} is the title of a debut song written by David Kent and Kirsti Manna, and performed by American country music artist [[Blake Shelton]]. \\
			\hline
		\end{tabular}		
		\caption{A sentence in the page ``Austin (song)''.}
		\label{e1}
	\end{table}
	
	In the above sentence, the mentions ``Austin'' and ``Blake Shelton'' in square brackets are annotated to the entity ``Austin (song)'' and ``Blake Shelton'', respectively. We generate candidates by searching sentences and thus obtain ``Blake Shelton'' as well as ``Austin (song)'' from this example. We reduce the number of candidates because many irrelevant entities linked by ``austin'' do not occur in returned sentences. In addition, as previous methods generate candidates by searching entities without the query information, ``austin'' can be linked to ``Austin, Texas'' with much higher static link-probability than all other senses of ``austin''. However, the number of returned sentences that contain ``Austin, Texas'' is close to the number of sentences that contain ``Austin (song)'' in our system. We show another example with the query ``her film'' in Table~\ref{e2}. In this sentence, ``Her'', ``romantic'', ``science fiction'', ``comedy-drama'' and ``Spike Jonze'' are annotated to corresponding entities. As ``Her'' is annotated to ``Her (film)'' by humans in this example, we have strong evidence to annotate it even if it is usually used as a possessive pronoun with very low static link-probability.
	
	\begin{table}[h]
		\centering
		\begin{tabular}{ |p{7cm}| }
			\hline			
			\textbf{[[Her (film)$\vert$Her]]} is a 2013 American [[romantic]] [[science fiction]] [[comedy-drama]] film written, directed, and produced by [[Spike Jonze]].\\
			\hline
		\end{tabular}		
		\caption{A sentence in the page ``Her (film)''.}
		\label{e2}
	\end{table}	
	
	We obtain the anchors as well as corresponding entities and map them to the query after searching similar sentences. Then we build a regression based framework to rank the candidates. We use a rich set of features, such as link-probability, context-matching, word embeddings, and relatedness among candidate entities as well as their related entities. We evaluate our method on the ERD14 and GERDAQ datasets. Experimental results show that our method outperforms state-of-the-art systems and yields 75.0\% and 56.9\% in terms of F1 metric on the ERD14 dataset and the GERDAQ dataset respectively.
	
	\section{Related Work}	

	Recognizing entity mentions in text and linking them to the corresponding entries helps to understand documents and queries. Most work uses the knowledge base including Freebase~\citep{chiu2014ntunlp}, YAGO~\citep{yosef2011aida} and Dbpedia~\citep{olieman2014entity}. Wikify~\citep{mihalcea2007wikify} is the very early work on linking anchor texts to Wikipedia pages. It extracts all n-grams that match Wikipedia concepts such as anchors and titles as candidates. They implement a voting scheme based on the knowledge-based and data-driven method to disambiguate candidates. \citet{cucerzan2007large} uses four recourses to generate candidates, namely entity pages, redirecting pages, disambiguation pages, and list pages. Then they disambiguate candidates by calculating the similarity between the contextual information and the document as well as category tags on Wikipedia pages.  \citet{milne2008learning} generate candidates by gathering all n-grams in the document, and retaining those whose probability exceeds a low threshold. Then they define commonness and relatedness on the hyper-link structure of Wikipedia to disambiguate candidates.
	
	The work on linking entities in queries has been extensively studied in recent years. 
	TagME~\citep{ferragina2010tagme} is a very early work on entity linking in queries. It generates candidates by searching Wikipedia page titles, anchors and redirects. Then disambiguation exploits the structure of the Wikipedia graph, according to a voting scheme based on a relatedness measure inspired by~\citet{milne2008learning}. The improved version of TagME, named WAT~\citep{piccinno2014tagme}, uses Jaccard-similarity between two pages' in-links as a measure of relatedness and uses PageRank to rank the candidate entities. Moreover, Meij~\citeyearpar{meij2012adding} proposes a two step approach for linking tweets to Wikipedia articles. They first extract candidate concepts for each n-gram, and then use a supervised learning algorithm to classify relevant concepts.
	 
	\begin{figure*}[]	
		\centering
		\includegraphics[width = 6.3in]{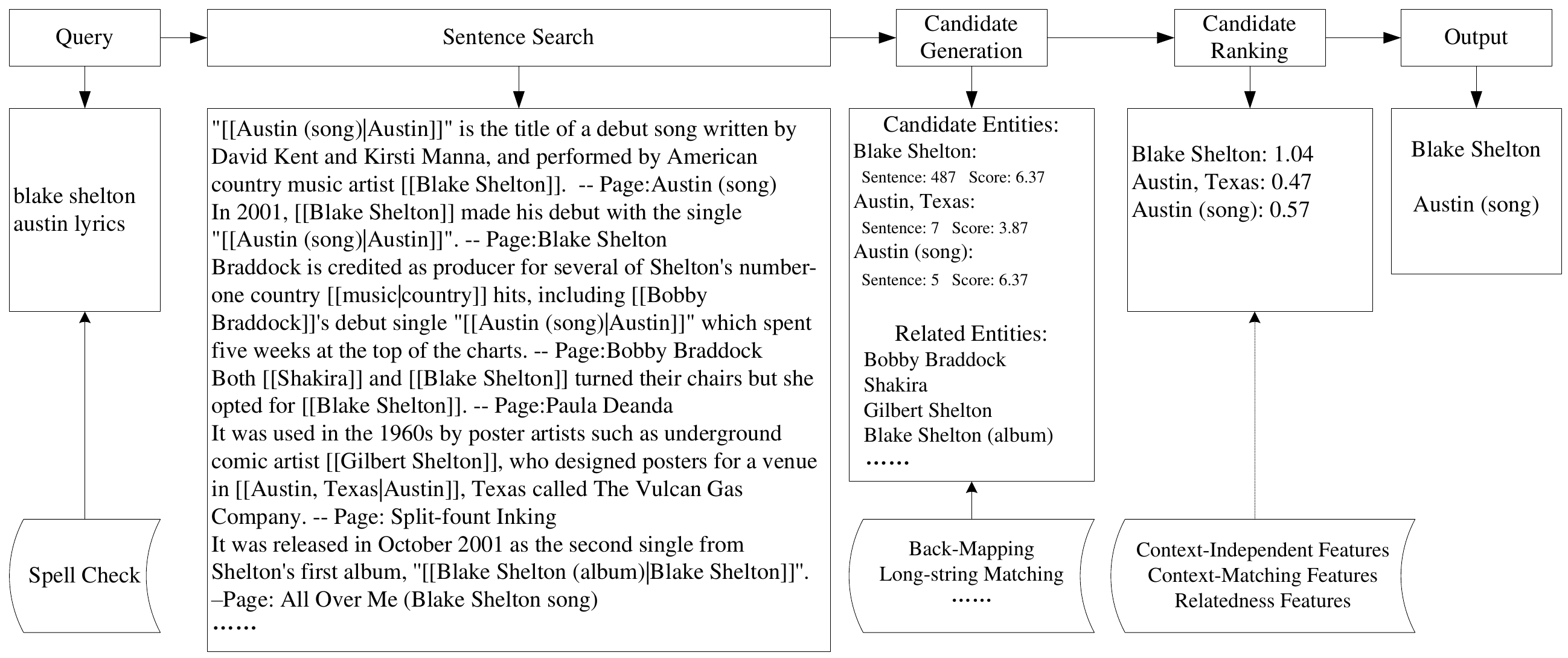}
		\caption{Example of the linking process of the query ``blake shelton austin lyrics''}
		\label{figure:austin}
	\end{figure*}	
	
	Unlike the work which revolves around ranking entities for query spans, the Entity Recognition and Disambiguation (ERD) Challenge~\citep{carmel2014erd} views entity linking in queries as the problem of finding multiple query interpretations. The SMAPH system~\citep{cornolti2014smaph} which wins the short-text track works in three phases: fetching, candidate-entity generation and pruning. First, they fetch the snippets returned by a commercial search engine. Next, snippets are parsed to identify candidate entities by looking at the boldfaced parts of the search snippets. Finally, they implement a binary classifier using a set of features such as the coherence and robustness of the annotation process and the ranking as well as composition of snippets. They further extend SMAPH-1 to SMAPH-2 \citep{cornolti2016piggyback}. They use the annotator WAT to annotate the snippets of search results to generate candidates and joint the additionally link-back step as well as the pruning step in the ranking phase, which gets the state-of-the-art results on the ERD14 dataset and their released dataset GERDAQ. There is another work closed to SMAPH that uses information of query logs and anchor texts~\citep{blanco2015fast}, which gives a ranked list of entities and is evaluated by means of typical ranking metrics. 			
	
	Our work is different from using search engines to generate candidates. We firstly propose to search Wikipedia sentences and take advantage of human annotations to generate candidates. The previous work, such as SMAPH, employs search engine for candidate generation, which puts queries in a	larger context in which it is easier to make sense of them. However, it uses WAT, an entity search based tool, to pre-annotate the snippets for candidate generation, which falls back the issues of entity search.	
	
	\section{Our Approach}
	
	As shown in Figure 1, we introduce our approach with the query ``blake shelton austin lyrics''. Our approach consists of three main phases: sentence search, candidate generation, and candidate ranking. First, we search the query in all Wikipedia articles to obtain the similar sentences. Second, we extract human-annotated entities from these sentences. We keep the entities whose corresponding anchor texts occur in the query as candidates, and treat others as related entities. Specifically, we obtain three candidates in this example, namely ``Blake Shelton'', ``Austin, Texas'', and ``Austin (song)''. Finally, we use a regression based model to rank the candidate entities. We get the final annotations of ``Blake Shelton'' and ``Austin (song)'' whose scores are higher than the threshold selected on the development set. In the following sections, we describe these three phases in detail.
	
	\subsection{Sentence Search}
	
	Sentences in Wikipedia articles usually contain anchors linking to entities. We are therefore motivated to generate the candidate entities based on the sentence search instead of the common method using entity search. There are some issues in the original annotations because of the annotation regulation. First, entities in their own pages are usually not annotated. Thus we annotate these entities with matching between the text and the page title. Second, entities are usually annotated only in their first appearance. We annotate these entities if they are annotated in previous sentences in the page. Moreover, pronouns are widely used in Wikipedia sentences and are usually not annotated. We use the Stanford CoreNLP toolkit \citep{manning-EtAl:2014:P14-5} to do the coreference resolution. In addition, we use the content in the disambiguation page and the infobox. Although these two kinds of information may have incomplete grammatical structure, it contains enough context information for the sentence search in our task.
	
	We use the Wikipedia snapshot of May 1, 2016, which contains 4.45 million pages and 120 million sentences. We extract sentences that contain at least one anchor in the Wikipedia articles, and extract human-annotated anchors as well as corresponding entities in the sentences. The original annotation contains 82.6 million anchors. We obtain 110 million annotated anchors in 48.4 million sentences after the incrementally annotation. All of above annotations are indexed by Lucene\footnote{\url{http://lucene.apache.org}} by building documents consisting of two fields: the first one contains the sentence and the second one contains all anchors with their corresponding entities. For each query, we search it with Lucene using its default ranker\footnote
	{Details can be found in \url{https://lucene.apache.org/core/2_9_4/api/core/org/apache/lucene/search/Similarity.html}} based on the vector space model and tf-idf to obtain the top K sentences (K is selected on the development set). We extract all entities as the related entities and use these sentences as their support sentences.

	\subsection{Candidate Generation}
	
	We back-map anchors and corresponding entities extracted in sentences to generate candidates. We use $(a, e)$ to denote the pair of the anchor text and corresponding entity and use $w(a, e)$ to denote the number of sentences containing the pair $(a, e)$. Then, we prune the candidate pairs according to following rules.
	
	First, we only keep the pair whose corresponding anchor text $a$ occurs in the query as a candidate, which has been used in previous work~\citep{ferragina2010tagme}.
	Second, we follow the long-string match strategy. If we have two pairs $(a_1,e_1)$ and $(a_2,e_2)$ while $a_1$ is a substring of $a_2$, we drop $(a_1,e_1)$ if $w(a_1,e_1) < w(a_2,e_2)$. This is because $a_2$ is typically less ambiguous than $a_1$. For example, for the query ``mesa community college football'', we can obtain the anchor ``mesa'', ``college'', ``community college'', and ``mesa community college''. We only keep ``mesa community college'' because it is longest and occurs most times in returned sentences. However, if $w(a_1,e_1) > w(a_2,e_2)$, we keep both candidate pairs because $a_1$ is more common in the query.
	
	In addition, we keep the entity whose surface form is the same with the anchor text and prune others. If we have two pairs $(a,e_1)$ and $(a,e_2)$ with the same anchor, and only $e_2$ occurs in the query, we drop the pair $(a,e_1)$ if $w(a,e_1) < w(a,e_2)$. For example, for the query ``business day south africa'', the anchor ``south africa'' can be linked to ``south africa'', ``union of south africa'', and ``south africa cricket team''. We only keep the entity ``south africa''.	
	
	\subsection{Candidate Ranking}
	
	We build a regression based framework to rank the candidate entities. In the training phase, we treat the candidates that are equal to the ground truth as the positive samples and the others as negative samples. The regression object of the positive sample is set to the score 1.0. The negative sample is set to the maximum score of overlapping ratio of tokens between its text and each gold answer. The regression object of the negative sample is not simply set to 0 in order to give a small score if the candidate is very closed to the ground truth. We find it benefits the final results. We use LIBLINEAR \citep{fan2008liblinear} with L2-regularized L2-loss support vector regression to train the regression model. The object function is to minimize		
	\begin{equation}
	w^Tw/2 + \mathnormal{C} \sum \max(0, |y_i-w^Tx_i|-eps)^2			
	\end{equation}
	where $x_i$ is the feature set, $y_i$ is the object score and $w$ is the parameter to be learned. We follow the default setting that $C$ is set to 1 and $eps$ is set to 0.1.	
	
	In the test phase, each candidate gets a score of $w^Tx_i$ and then we only output the candidate whose score is higher than the threshold selected on the development set.
	
	We employ four different feature sets to capture the quality of a candidate from different aspects. All features are shown in Table \ref{table:Feature}. 
	\begin{description}[font=\textbf, labelindent=0em, leftmargin=0em, style=sameline]
		\item [Context-Independent Features]
		
		This feature set measures each annotation pair $(a,e)$ without context information. Feature 1-4 catch the syntactic properties of the candidate. Feature 5 is the number of returned sentences that contain $(a,e)$. Feature 6 is the maximum search score (returned by Lucene) in its support sentences. Moreover, inspired by TagME~\citep{ferragina2010tagme}, we denote $freq(a)$ as the number of times the text $a$ occurs in Wikipedia. We use $link(a)$ to denote the number of times the text $a$ occurs as an anchor. We use $lp(a) = link(a)/freq(a)$ to denote the static link-probability that an occurrence of $a$ has been set as an anchor. We use $freq(a,e)$ to denote the number of times that the anchor text $a$ links to the entity $e$, and use $pr(e|a)$ = $freq(a,e)/link(a)$ to denote the static prior-probability that the anchor text $a$ links to $e$. Features 7 and 8 are these two probabilities.
		
		\item [Context-Matching Features]
		
		\begin{table}[t]
			\small	
			\centering
			\begin{tabular}{p{0.2cm} | p{1.25cm} |p{4.9cm} }
				\hline
				ID	& Name & Description \\
				\hline
				1 & $in\_query$ & 1 if $e$ is in the query, 0 otherwise \\	
				2 & $is\_pt$ & 1 if $e$ contains parenthesis, 0 otherwise\\	
				3 & $is\_cm$ & 1 if $e$ contains comma, 0 otherwise\\	
				4 & $len$ & len($e$) by tokens \\	
				5 & $w(a,e)$ & number of support sentences \\	
				6 & $sc(a,e)$ & maximum search score of support sentences \\	
				7 & $lp(a)$ & static link-probability that $a$ is an anchor \\	
				8 & $pr(a,e)$ & static prior-probability that $a$ links to $e$ \\	
				9 & $cm\_sc$ & context matching score to the support sentences \\	
				10 & $cm\_fs$ & context matching score to the first sentence of $e$'s page \\	
				11 & $cm\_dd$ & context matching score to the description in $e$'s disambiguation page \\	
				12 & $embed\_sc$ & maximum embedding similarity of the query and each support sentence \\	
				13 & $embed\_fs$ & embedding similarity of the query and the first sentence of $e$'s page \\	
				14 & $embed\_dd$ & embedding similarity of the query and the description in $e$'s disambiguation page \\
				15 & $rel\_cd\_sc$ & number of candidates that occur in the support sentences \\
				16 & $rel\_cd\_sp$ & number of candidates that occur in the same Wikipedia page \\
				17 & $rel\_re\_sc$ & number of related entities that occur in the support sentences\\ 
				18 & $rel\_re\_sp$ & number of related entities that occur in the same Wikipedia page \\
				\hline
			\end{tabular}
			\caption{Feature Set for Candidate Ranking}
			\label{table:Feature}
		\end{table}
		
		We treat the other words except for the anchor text as the context. This feature set measures the context matching to the query. Feature 9 is the context matching score calculated by tokens. We denote $c$ as the set of context words. For each $c_i$ in $c$, the $cm\_sc(c_i)$ is the ratio of times that $c_i$ occurs in the support sentences, and $cm\_sc(c) = \frac{1}{N}\sum{cm\_sc(c_i)}$. Features 10 and 11 are the ratio of context words occurring in the first sentence in the entity page and the description of entity's disambiguation page (if existed), respectively. Moreover, we train a 300-dimensional word embeddings on all Wikipedia articles by word2vec~\citep{mikolov2013efficient} and use the average embedding of each word as the sentence representation. Feature 12 is the maximum cosine score between the query and each support sentence. Features 13 and 14 are calculated with the first sentence in the entity's page and the description in the disambiguation page. 
		
		\item [Relatedness Features of Candidate Entities]
		This set of features measures how much an entity is supported by other candidates. Feature 15 is the number of other candidate entities occurring in the support sentences. Feature 16 is the number of candidate entities occurring in the same Wikipedia page with the current entity.
		
		\item [Relatedness Features to Related Entities]
		This set of features measures the relatedness between candidates and related entities outside of queries. Related entities can provide useful signals for disambiguating the candidates. Features 17 and 18 are analogous features with features 15 and 16, which are calculated by the related entities.
	\end{description}
	
	\section{Experiment}
	
	We conduct experiments on the ERD14 and GERDAQ datasets. We compare with several baseline annotators and experimental results show that our method outperforms the baseline on these two datasets. We also report the parameter selection on each dataset and analyze the quality of the candidates using different methods.
	
	\subsection{Dataset}
	\begin{description}[font=\textbf, labelindent=0em, leftmargin=0em, style=sameline]
		\item [ERD14\footnote{\url{http://web-ngram.research.microsoft.com/erd2014/Datasets.aspx}}] is a benchmark dataset in the ERD Challenge~\citep{carmel2014erd}, which contains both long-text track and short-text track. In this paper we only focus on the short-text track. It contains 500 queries as the development set and 500 queries as the test set. Due to the lack of training set, we use the development set to do the model training and tuning. This dataset can be evaluated by both Freebase and Wikipedia as the ERD Challenge Organizers provide the Freebase Wikipedia Mapping with one-to-one correspondence of entities between two knowledge bases. We use Wikipedia to evaluate our results.
				
		\item [GERDAQ\footnote{\url{http://acube.di.unipi.it/datasets}}] is a benchmark dataset to annotate entities to Wikipedia built by~\citet{cornolti2016piggyback}. It contains 500 queries for training, 250 for development, and 250 for test. The query in this dataset is sampled from the KDD-Cup 2005 and then annotated manually. Both name entities and common concepts are annotated in this dataset.
	\end{description}
	
	\subsection{Evaluation Metric}
	We use average F1 designed by ERD Challenge~\citep{carmel2014erd} as the evaluation metrics. Specifically, given a query q, with labeled entities $\hat{A}=\{\hat{E_1},\dots,\hat{E_n}\}$. We define the F-measure of a set of hypothesized interpretations $A = \{E_1,\dots,E_m\}$ as follows:
	\begin{align}
	&Precision = \frac{\vert\hat{A} \cap A\vert}{\vert A \vert}, Recall = \frac{\vert\hat{A} \cap A\vert}{\vert \hat{A} \vert}	\\
	&F_1 = \frac{2 \times Precision \times Recall}{Precision + Recall}
	\end{align}
	The average F1 of the evaluation set is the average of the F1 for each query:
	\begin{equation}
	Average F_1 = \frac{1}{N}\sum_{i=1}^{N}F_1(q_i)
	\end{equation}	
	Following the evaluation guideline in ERD14 and GERDAQ, we define recall to be 1.0 if the gold binding of a query is empty and define precision to be 1.0 if the hypothesized interpretation is empty.
	
	\subsection{Baseline Methods}
	
	We compare with several baselines and use the results reported by the ERD organizer and~\citet{cornolti2016piggyback}.
	
	\begin{description}[font=\textbf, labelindent=0em, leftmargin=0em, style=sameline,noitemsep]
		\item [AIDA]~\citep{hoffart2011robust} searches the mention using Stanford NER Tagger based on YAGO2. We select AIDA as a representative system aiming to entity linking for documents following the work in \citet{cornolti2016piggyback}.
		\item [WAT]~\citep{piccinno2014tagme} is the improved version of TagME~\citep{ferragina2010tagme}. 
		\item [Magnetic IISAS]~\citep{laclavik2014search} retrieves the index extracted from Wikipedia, Freebase and Dbpedia. Then it exploits Wikipedia link graph to assess the similarity of candidate entities for disambiguation and filtering.
		\item [Seznam]~\citep{eckhardt2014entity} uses Wikipedia and DBpedia to generate candidates. The disambiguation step is based on PageRank over the graph.
		\item [NTUNLP]~\citep{chiu2014ntunlp} searches the query to match Freebase surface forms. The disambiguation step is built on top of TagME and Wikipedia.
		\item [SMAPH-1]~\citep{cornolti2014smaph} is the winner in the short-text track in the ERD14 Challenge. 
		\item [SMAPH-2]~\citep{cornolti2016piggyback} is the improved version of SMAPH-1. It generates candidates from the snippets of search results returned by the Bing search engine. 
		
	\end{description}
	
	\begin{table}[h]	
	\small
		\centering
		\begin{tabular}{l | c }
			\hline
			System	& $F1_{avg}$ \\
			\hline
			AIDA & 22.1 \\			
			WAT & 58.6 \\	
			Magnetic IISAS& 65.6 \\ 
			Seznam & 66.9 \\
			NTUNLP &68.0\\
			SMAPH-1 & 68.8\\ 	
			SMAPH-2 & 70.8 \\
			\hline
			Our work & \textbf{75.0*}\\
			w/o Spell Check & 74.0\\
			w/o Additional Annotation & 74.4 \\
			w/o Context Feature & 72.6 \\
			w/o Relatedness Feature & 74.5\\
			\hline
		\end{tabular}
		\caption{Results on the ERD dataset. Results of the baseline systems are taken from Table 8 in~\citet{cornolti2016piggyback} and reported by the ERD organizer~\citep{carmel2014erd}. We only report the F1 score as precision and recall are not reported in previous work. *Significant improvement over state-of-the-art baselines (t-test, p $<$ 0.05).
		}
		\label{table:ERD_Reslut}
		
		\vspace{0.397cm}

		\begin{tabular}{l | p{0.6cm} | p{0.6cm}| p{0.8cm} }
			\hline
			System	& $P_{avg}$	& $R_{avg}$	& $F1_{avg}$ \\
			\hline
			AIDA & 94.0 & 12.2 & 12.6 \\			
			TagME & 60.4 & 51.2 & 44.7 \\
			WAT & 49.6 & 57.0 & 46.0 \\	
			SMAPH-1 & 77.4 & 54.3& 52.1 \\
			SMAPH-2 & 72.1 & 55.3 & 54.4 \\
			\hline
			Our work &71.5 & 58.5 &\textbf{56.9}\\  
			w/o Spell Check &75.4 & 48.6& 49.3\\
			w/o Additional Annotation &70.3 & 58.2 & 55.8 \\
			w/o Context Feature &69.2& 56.4 & 55.5\\
			w/o Relatedness Feature &73.3& 57.4& 56.7 \\	
			\hline	
		\end{tabular}
		\caption{Results on the GERDAQ dataset. Results of the baseline systems are taken from Table 10 in~\citet{cornolti2016piggyback}.}
		\label{table:GERDAQ_Reslut}
	\end{table}
	\subsection{Result}
	
	We report results on the ERD datset and GERDAQ dataset in Table~\ref{table:ERD_Reslut} and Table~\ref{table:GERDAQ_Reslut}, respectively. On the ERD14 dataset, WAT is superior to AIDA but it is still up to $10\%$ than SMAPH-1 that wins the ERD Challenge. SMAPH-2 improves $2\%$ than SMAPH-1. Our system significantly outperforms the state-of-the-art annotator SMAPH-2 by $4.2\%$. On the GERDAQ dataset, our system is $2.5\%$ superior to the state-of-the-art annotator SMAPH-2. The F1 score in this dataset is much lower than the ERD dataset because common concepts such as ``Week'' and ``Game'' that are not annotated in the ERD dataset are annotated in the GERDAQ dataset.
	
	Spell checking has been widely used in the baseline annotators as it is not uncommon in queries~\citep{laclavik2014search}. The SMAPH system that generates candidates by search results implicitly leverages the spell-checking embedded in search engines. In our experiments, spell checking improves 1.0\% on the ERD dataset and 7.6\% on the GERDAQ dataset. Furthermore, only $6.9\%$ of queries in the ERD14 dataset have spelling mistakes, whereas the number in the GERDAQ dataset is $23.0\%$. Thus spell-checking is more important in the GERDAQ dataset. 
	
	The result decreases 0.6\% on the ERD dataset and 1.1\% on the GERDAQ dataset without the additional annotation. Furthermore, while the F1 score decreases 2.4\% on the ERD dataset and 1.4\% on the GERDAQ dataset without the context features, the score only decreases 0.5\% on the ERD dataset and 0.2\% on the GERDAQ dataset without the relatedness features. Unlike the work on entity linking for documents~\citep{eckhardt2014entity,witten2008effective} that features derived from entity relations get promising results, the context features play a more important role than the relatedness features on entity linking for queries as search queries are short and contain fewer entities than documents.

	\begin{figure}
		\centering
		\includegraphics[width=2.94in]{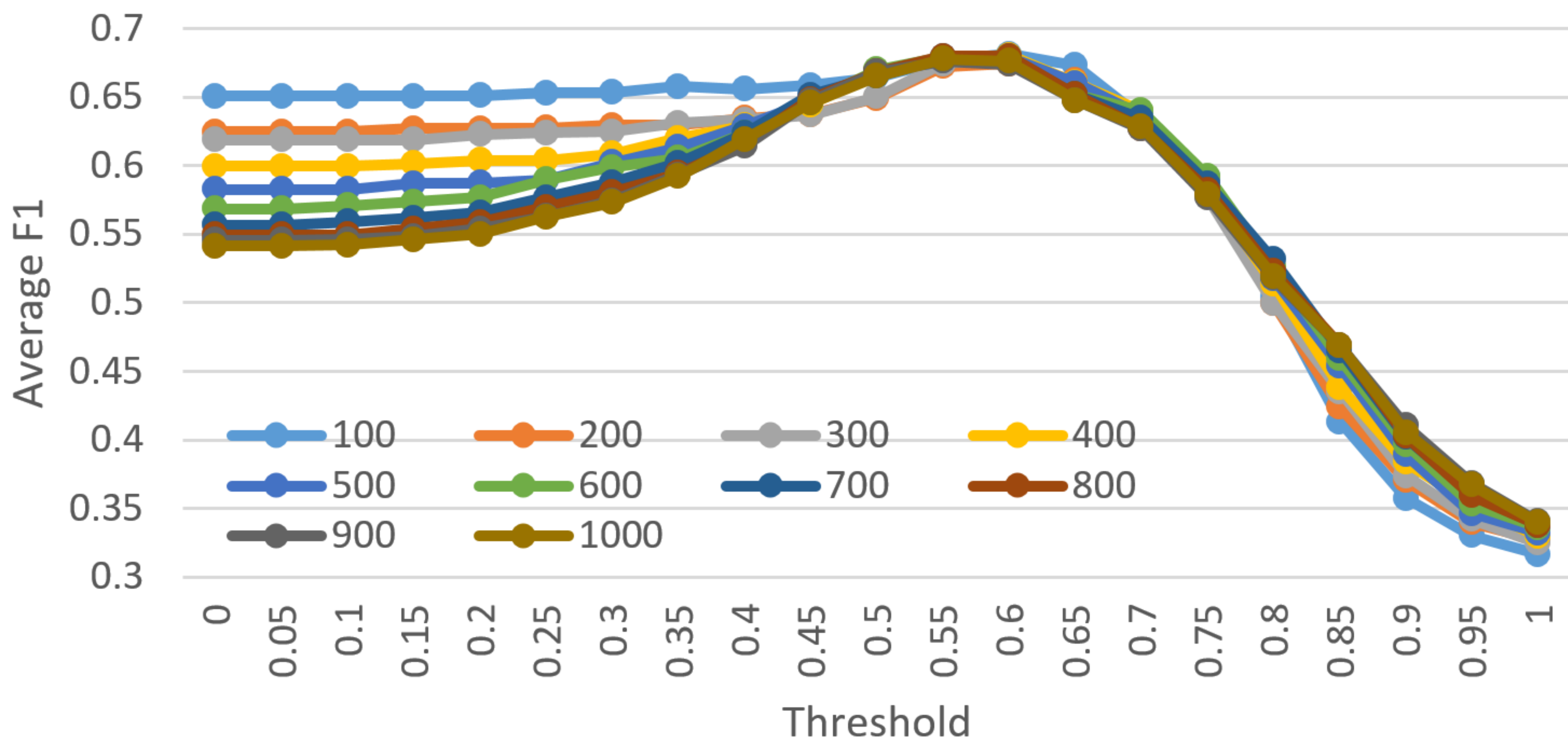}
		\caption{F1 scores with different search numbers and thresholds on the ERD development set}	
		\label{fig:para_erd} 	
		
		\centering
		\includegraphics[width=2.94in]{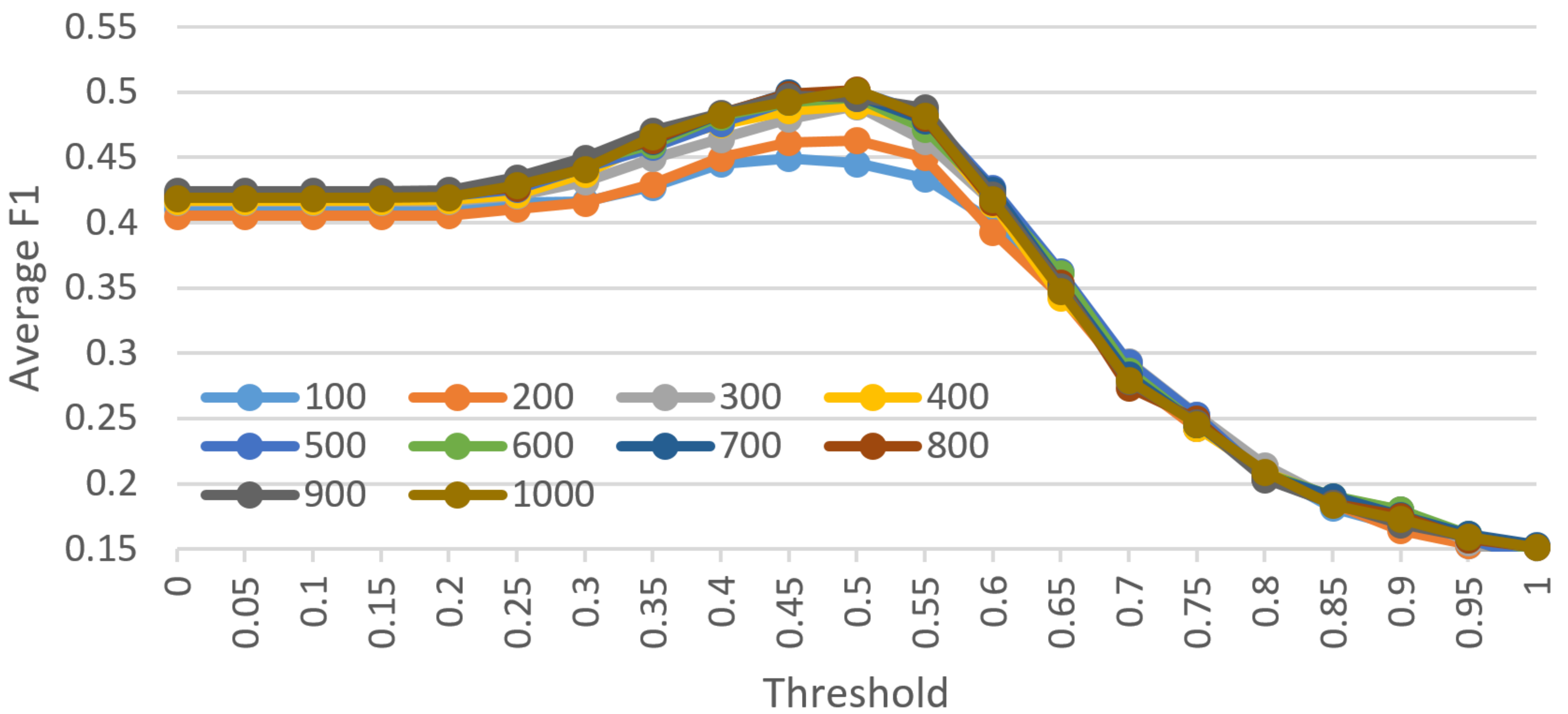}
		\caption{F1 scores with different search numbers and thresholds on the GERDAQ development set}	
		\label{fig:para_gerdaq}
	\end{figure}
	\subsection{Parameter Selection}
	
	There are two parameters in our framework, namely the number of search sentences and the threshold for final output. We select these two parameters on the development set. We show the F1 score with different numbers of search sentences and thresholds in Figure~\ref{fig:para_erd} and Figure~\ref{fig:para_gerdaq}. On the ERD development set, better results occur in the search number between 600 and 800 as well as the threshold 0.55 and 0.6. On the GERDAQ development set, better results occur in the search number between 700 and 1000 as well as the threshold between 0.45 and 0.5. In our experiment, we set the number of sentences to 700 and the threshold to 0.56 on the ERD dataset as well as 800 and 0.48 on the GERDAQ dataset according to the F1 scores on the development set.
	\begin{table}
		\small
		\centering
		\begin{tabular}{p{2.2cm}| c |c|p{0.8cm}}
			\hline
			\multirow{2}{*}{Method}	& Number of & Number of 
			& \multirow{2}{*}{$F1_{avg}$}  \\
			& anchors &candidates &\\		
			\hline
			Entity Search & \multirow{2}{*}{1.96} & \multirow{2}{*}{6.84} & 66.46 \\
			ES + RF & &  &69.00\\
			\hline
			Sentence Search & \multirow{2}{*}{1.12} & \multirow{2}{*}{1.49} & 73.81 \\				
			SS + RF  &  & & 75.01\\
			\hline		
		\end{tabular}
		\caption{Comparison with different candidate generation methods on the ERD dataset. +RF: integrating ranking features extracted by Sentence Search.}
		\label{table:number_of_candidates}
	\end{table}
	
	\begin{table}
		\centering
		\small		
		\begin{tabular}{c|c|c|c}
			\hline
			Method & $C_{avg}$ & $P_{avg}$ & $R_{avg}$ \\
			\hline		
			Entity Search  & 78.87 & 77.56 & 66.04 \\				
			Sentence Search & 74.42 & 89.61 & 69.08 \\
			\hline		
		\end{tabular}
		\caption{Results for the 398 queries which have at least one labeled entity on the ERD dataset using different candidate generation methods. $C_{avg}$ is the average recall of candidates per query. $P_{avg}$ and $R_{avg}$ are calculated on the final results.}
		\label{table:coverage_threshold}
	\end{table}	
	
	\subsection{Model Analysis} 
	
	The main difference between our method and most previous work is that we generate candidates by searching Wikipedia sentences instead of searching entities. For generating candidates with entity search, we build a dictionary containing all anchors, titles, and redirects in Wikipedia. Then we query the dictionary to get the mention and obtain corresponding entities as candidates. We use the same pruning rules and ranking framework in our experiments, but exclude the features from support sentences because the entity search method does not contain the information. The F1 score is shown in Table~\ref{table:number_of_candidates}. We achieve similar results in our implementation of the method using entity search on the ERD dataset as Magnetic IISAS~\citep{laclavik2014search} which uses a similar method and ranks 4th with the F1 of 65.57 in the ERD14 Challenge. 
	
	We compare the two candidate generation methods in several aspects. First, we show the overall results in Table~\ref{table:number_of_candidates}. The average number of candidates from our method is much smaller. It is noted that the anchors from sentence search can also be found in entity search. However, we only extract the entities in the returned sentences while the methods by entity search use all entities linked by the anchors. In addition, features such as the number of sentences containing the entity from sentence search which provide query sensitive prior probability contribute to the ranking process. It improves the F1 score from 73.81 to 75.01 for sentence search and from 66.46 to 69.00 for entity search. More important, the result of ``ES+RF'' is still significantly worse than the result of both small candidate set and Wikipedia related features that prunes irrelevant candidates at the beginning, which proves that the high-quality candidate set is very important since the larger candidate set brings in lots of noise in training a ranking model. Moreover, there are 102 queries (20.4\%) without labeled entities in the ERD dataset. We only give 7 incorrect annotations in these queries while the number is 13 from entity search. Furthermore, as shown in Table~\ref{table:coverage_threshold}, the coverage of our method is lower in queries with at least one entity, but we obtain better results on precision, recall and F1 in the final stage. 
	
	\begin{figure}
		\centering
		\includegraphics[width=3in]{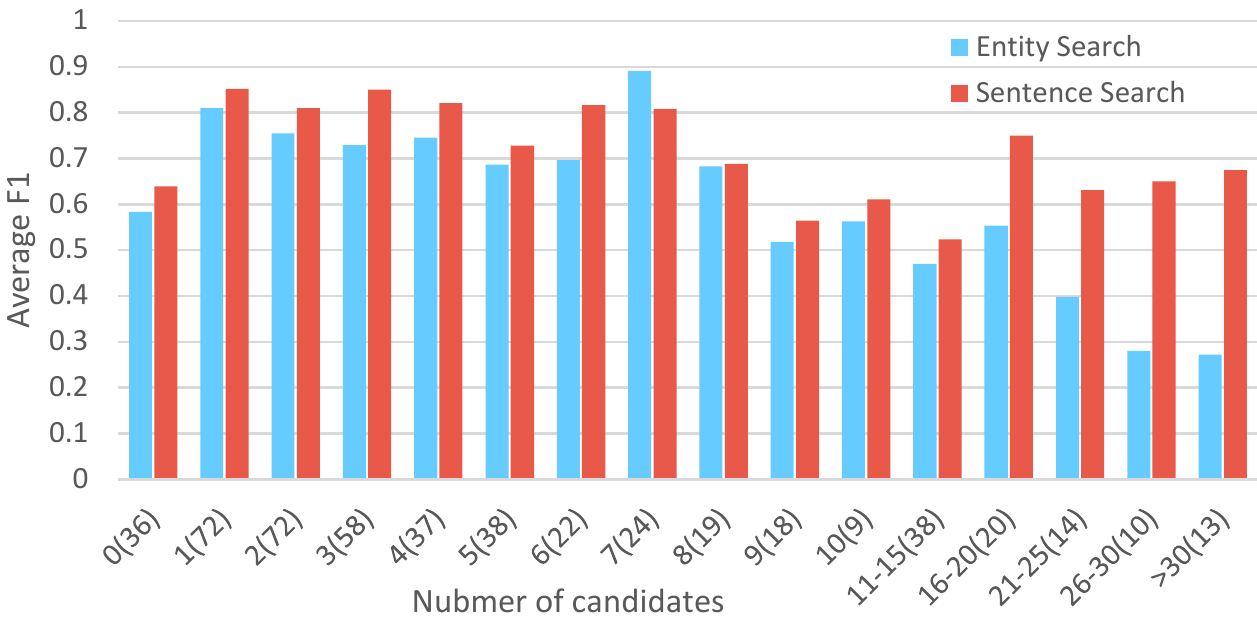}
		\caption{F1 scores with number of candidates using different methods on the ERD dataset. The number of queries is shown in the parentheses.}	
		\label{fig:analysis} 			
	\end{figure}	
	
	Figure~\ref{fig:analysis} illustrates the F1 score grouped by the number of candidates using entity search. In almost all columns the F1 score of our method is better than the baseline. In left columns (the number of candidates is less than 10), both methods generate few candidates. The F1 score of our method is higher, which proves that we train a better ranking model because of our small but quality candidate set. Moreover, the right columns (the number of candidates is more than 10) show that the F1 score using entity search gradually decreases with the incremental candidates. However, our method based on sentence search takes advantage of context information to keep a small set of candidates, which keeps a consistent result and outperforms the baseline.
	
	\section{Conclusion}
	
	In this paper we address the problem of entity linking for open-domain queries. We introduce a novel approach to generating candidate entities by searching sentences in the Wikipedia to the query, then we extract the human-annotated entities as the candidates. We implement a regression model to rank these candidates for the final output. Two experiments on the ERD dataset and the GERDAQ dataset show that our approach outperforms the baseline systems. In this work we directly use the default ranker in Lucene for similar sentences, which can be improved in future work.
	
	\section*{Acknowledgments}
	We thank Ming-Wei Chang for sharing the ERD14 dataset. The first author and the fourth author are supported by the National Natural Science Foundation of China (Grant No. 61421003). 

\bibliography{emnlp2017}
\bibliographystyle{emnlp_natbib}

\end{document}